\documentclass[a4paper,fleqn]{cas-dc}

\makeatletter
\ExplSyntaxOn
\cs_set_eq:NN \__first_footerline: \prg_do_nothing:
\ExplSyntaxOff
\makeatother

\usepackage[numbers]{natbib}
\usepackage{tabularx}
\usepackage{fancyvrb}
\usepackage{fvextra}

\def\tsc#1{\csdef{#1}{\textsc{\lowercase{#1}}\xspace}}
\tsc{WGM}
\tsc{QE}
\tsc{EP}
\tsc{PMS}
\tsc{BEC}
\tsc{DE}
\begin{document}
\let\WriteBookmarks\relax
\def\floatpagepagefraction{1}
\def\textpagefraction{.001}
\shorttitle{}
\shortauthors{T. Qin et~al.}

\title [mode = title]{A Physical Testbed for Self-Driving Laboratories: Comparing Bayesian and Agentic Optimization in a Minimal Closed-Loop Color Formulation Task}                      

\author[1,2]{Tony Qin}%[type=editor,
                       % auid=000,bioid=1,
                        %orcid=0000-0001-0000-0000]
% \ead{tonyqin@cs.unc.edu}

\credit{Conceptualization, Data curation, Investigation, Methodology, Software, Visualization, Writing -- original draft, Writing -- review \& editing}

\affiliation[1]{organization={Department of Computer Science, The University of North Carolina at Chapel Hill},
                city={Chapel Hill},
%               citysep={}, % Uncomment if no comma needed between city and postcode
                state={North Carolina},
                postcode={27599}, 
                country={USA}}

\author[2]{Ziheng Wu}
\credit{Writing -- original draft, Writing -- review \& editing}

\affiliation[2]{organization={Lawrence Livermore National Laboratory},
                city={Livermore},
                state={California},
                postcode={94550},
                country={USA}}

\author[3]{Maximilian Schommer}
\credit{Software}

\author[3]{Roya Amini-Naieni}
\credit{Conceptualization, Visualization, Writing -- review \& editing}

\author[2]{Aldair Gongora}
\cormark[1]
\ead{gongora1@llnl.gov}
\credit{Conceptualization, Data curation, Investigation, Supervision, Visualization, Writing -- original draft, Writing -- review \& editing}

\affiliation[3]{organization={Trilobio},
                city={San Francisco},
                state={California}, 
                postcode={94107}, 
                country={USA}}

\cortext[cor1]{Corresponding author}

\begin{abstract}
Self-driving laboratories (SDLs) integrate laboratory robotics and automation with AI-driven decision-making to close the design–build–test–learn (DBTL) cycle. However, physical testbeds that combine off-the-shelf hardware and AI into a functioning closed-loop workflow remain in early stages of development, and their integration with commercial laboratory automation is comparatively less developed. In this work, we develop a physical testbed that assembles and integrates commercially available laboratory automation instruments and demonstrate continuous, closed-loop, autonomous operation over 24 cumulative hours of unattended experimentation. We validate this platform on a minimal closed-loop discovery task in which the system autonomously formulates aqueous dye mixtures to match a target absorbance spectrum. The system completed six independent campaigns, each consisting of 48 closed-loop iterations over approximately four hours. In every campaign, the system autonomously selected, dispensed, and measured each mixture, with an AI-driven decision-making algorithm determining subsequent formulations without human intervention. Using this testbed, we compared the performance of two AI-driven decision-making algorithms, traditional Bayesian optimization (BO) and an agentic approach, in selecting mixtures under a fixed total-volume constraint. The agentic optimizer matched target absorbance spectra using fewer iterations than BO, with reductions up to 52\%, and converged on a solution for one target that BO failed to match within 48 iterations. These results demonstrate a reference physical platform for SDLs that can evaluate both established and emerging AI strategies, including large language model (LLM)-directed optimization, before their deployment in real-world applications where experiments may be costly and time-consuming to prepare.
\end{abstract}

\begin{keywords}
self-driving laboratory\sep Bayesian optimization\sep agentic optimization\sep color formulation
\end{keywords}

\maketitle

\section{Introduction}
Self-driving laboratories (SDLs) integrate machine learning, robotics, and automated instrumentation to close the design-build-test-learn (DBTL) loop, in which a decision-making algorithm proposes an experiment, an automated platform executes it, and the resulting measurements inform subsequent experiments \cite{Burger2020_Mobile, MacLeod2020_Selfdriving, Stach2021_Autonomous, Abolhasani2023_Rise, Tom2024_SelfDriving, Canty2026_Present}. Their foundations extend back to the automation of physical workflows in laboratory settings \cite{Skeggs1957_Automatic, Olsen2012_First}, while later work integrated optimization algorithms to automate the decision-making process \cite{Parker1978_Computerassisted}. A subsequent milestone was the Robot Scientist work of King et al., which demonstrated automated cycles of hypothesis generation and experimentation \cite{King2004_Functional, King2009_Automation}. These studies established a foundation for integrating scientific reasoning with physical experimentation.

Building on this, a prominent wave of demonstrations expanded autonomous experimentation across chemistry and materials science. Examples include a mobile robotic chemist for photocatalyst formulation \cite{Burger2020_Mobile}, a modular platform for optimizing thin-film materials \cite{MacLeod2020_Selfdriving}, and the Bayesian experimental autonomous researcher (BEAR) for mechanical design \cite{Gongora2020_Bayesian}. The subsequent A-Lab integrated robotic powder handling, synthesis, and characterization with computational and literature-derived information to guide inorganic materials synthesis \cite{Szymanski2023_Autonomous}. Together with the broader landscape of SDLs \cite{Tom2024_SelfDriving,Canty2026_Present, Angelopoulos2024_Transforming}, these platforms illustrate the growing range of scientific problems accessible to closed-loop experimentation.

\begin{figure*}
    \centering
    \includegraphics[width=\textwidth]{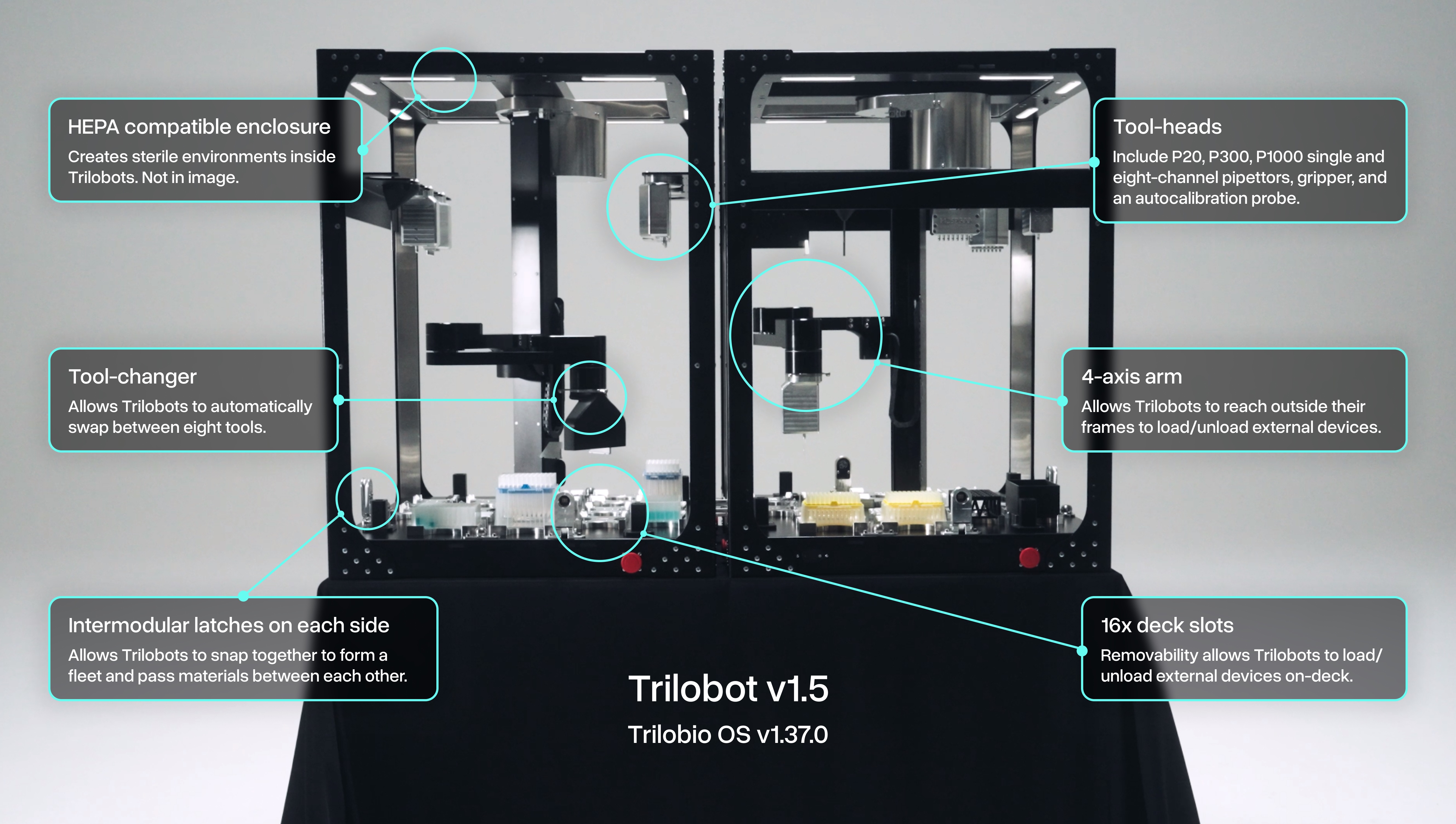}
    \caption{Two Trilobot robots placed side by side to display their components. The Trilobot's 4-axis arm has a tool changer, allowing it to swap between various end effector tools, including a pipette handler and grippers. The base of the platform has deck slots where equipment such as microplates and pipette tip racks can be placed.}
    \label{fig:trilobot}
\end{figure*} 
Color formulation has emerged as a recurring SDL benchmark \cite{Roch2020_ChemOS, Ginsburg2023_Exploring, Delley2026_Semantic} as it enables low-cost experimentation for alternative decision making and automation procedures. This problem captures fundamental components of autonomous discovery, including experiment proposal, sample creation, and sample analysis \cite{Ginsburg2023_Exploring}. Furthermore, liquid-based color formulation requires liquid-handling procedures, which are common to many biology and chemistry experiments \cite{Wierenga2023_PyLabRobot}. As a result, color formulation can serve as an SDL reference problem, making it suitable for evaluating general-purpose decision-making methods, while its interpretable structure and potential analytical baselines facilitate validation. Prior demonstrations span closed-loop spectral light mixing \cite{Baird2023_Building, Kitchin2025_Evolving} and pigment-based color matching \cite{Ginsburg2023_Exploring, Yoshikawa2026_NIMO}.

The concept of a ``minimal working example'' distills an SDL into its essential operations, such as issuing commands, acquiring measurements, and using an optimization algorithm to select the next experiment \cite{Baird2022_What}. In doing so, it allows the exploration of autonomous laboratories as physical testbeds in low-risk environments. The Closed-loop Spectroscopy Lab Light-mixing Demo by Baird et al. demonstrated this concept, mixing RGB LED light and reading the resulting spectrum with a light sensor to match a target color \cite{Baird2023_Building}. While concepts such as minimal working examples or ``frugal'' SDLs have been proposed \cite{Lo2024_Review}, few studies have explored the integration of commercial laboratory automation specifically for investigating SDL technology development. Existing studies that incorporate commercial equipment in closed-loop platforms tend to focus primarily on domain-specific scientific challenges \cite{Burger2020_Mobile, MacLeod2020_Selfdriving}, rather than on the underlying principles of SDL operation in low-risk, minimal-complexity scenarios.

Rapid advances in large language models (LLMs) \cite{Sajadieh2026_Artificial} have further expanded the possibilities for scientific automation. In SDLs, LLMs can support workflow orchestration \cite{Delley2026_Semantic, Angelopoulos2026_Prompts}, literature and protocol interpretation, hypothesis generation, computational tool use, data analysis, anomaly diagnosis, and human–machine interaction \cite{M.Bran2024_Augmenting, Muller2026_Agente}. For example, Coscientist demonstrated how an LLM-based system could combine information retrieval, code execution, and laboratory tools to plan and perform chemical experiments \cite{Boiko2023_Autonomous}. These capabilities offer routes for translating scientific objectives into executable workflows and adapting those workflows as experimental requirements evolve.

LLMs can also participate directly in experimental decision-making by proposing candidate conditions from the objective and accumulated observations \cite{Muller2026_Agente, Yang2024_Large, Liu2024_Large, CaldasRamos2026_Bayesian}. Although decision-making in SDLs is traditionally performed with a method such as Bayesian optimization (BO) \cite{Tom2024_SelfDriving}, an LLM agent can perform this step by writing the code for the optimizer and monitoring its execution. These developments motivate evaluating both conventional BO and LLM-directed candidate selection within a common physical workflow.

In this work, we address the challenge of evaluating decision-making approaches within a common physical testbed built from off-the-shelf laboratory instrumentation. We present a platform that integrates a commercially available laboratory robot with commercially sourced measurement instrumentation into a single autonomous, closed-loop system. Developing the testbed from commercial hardware reduces variability arising from custom implementations and enables evaluation under consistent conditions. The platform combines a modular laboratory robot (Trilobot, Trilobio), an absorbance measurement instrument (Absorbance 96 Automate, Byonoy), and an interchangeable optimizer to formulate mixtures of red, yellow, and blue dye solutions and water under a fixed total-volume constraint. These components form an integrated hardware and software system in which every iteration, from liquid transfer and microplate handling to spectral measurement, model update, and selection of the next formulation, proceeds without human intervention. 

We evaluated the testbed across six independent campaigns; the system completed 48 closed-loop iterations per campaign over approximately four hours without manual intervention. The system reproduced target absorbance spectra with an average root mean squared error (RMSE) of $9.95 \times 10^{-3}$ optical density (OD), matching the target spectrum within measurement error in five out of six campaigns. We further compared BO with an LLM-based decision-making agent within the same experimental loop, and the agentic optimizer matched all three target spectra within fewer iterations. The underlying operations, liquid handling and spectral measurement performed by the hardware, together with the automated decision-making that governs optimization, are common across many experimental workflows in materials science, chemistry, and biology. However, we note that while representative, this workflow does not establish generalizability across the full range of experimental techniques or material systems used in these fields. Ultimately, these results underscore the role of physical testbeds in advancing SDL technology, providing a proving ground in which both hardware instrumentation and decision-making strategies can be evaluated prior to deployment in real-world scientific processes.

\section{Developing an Autonomous Laboratory}
\begin{figure}
    \centering
    \includegraphics[width=\linewidth]{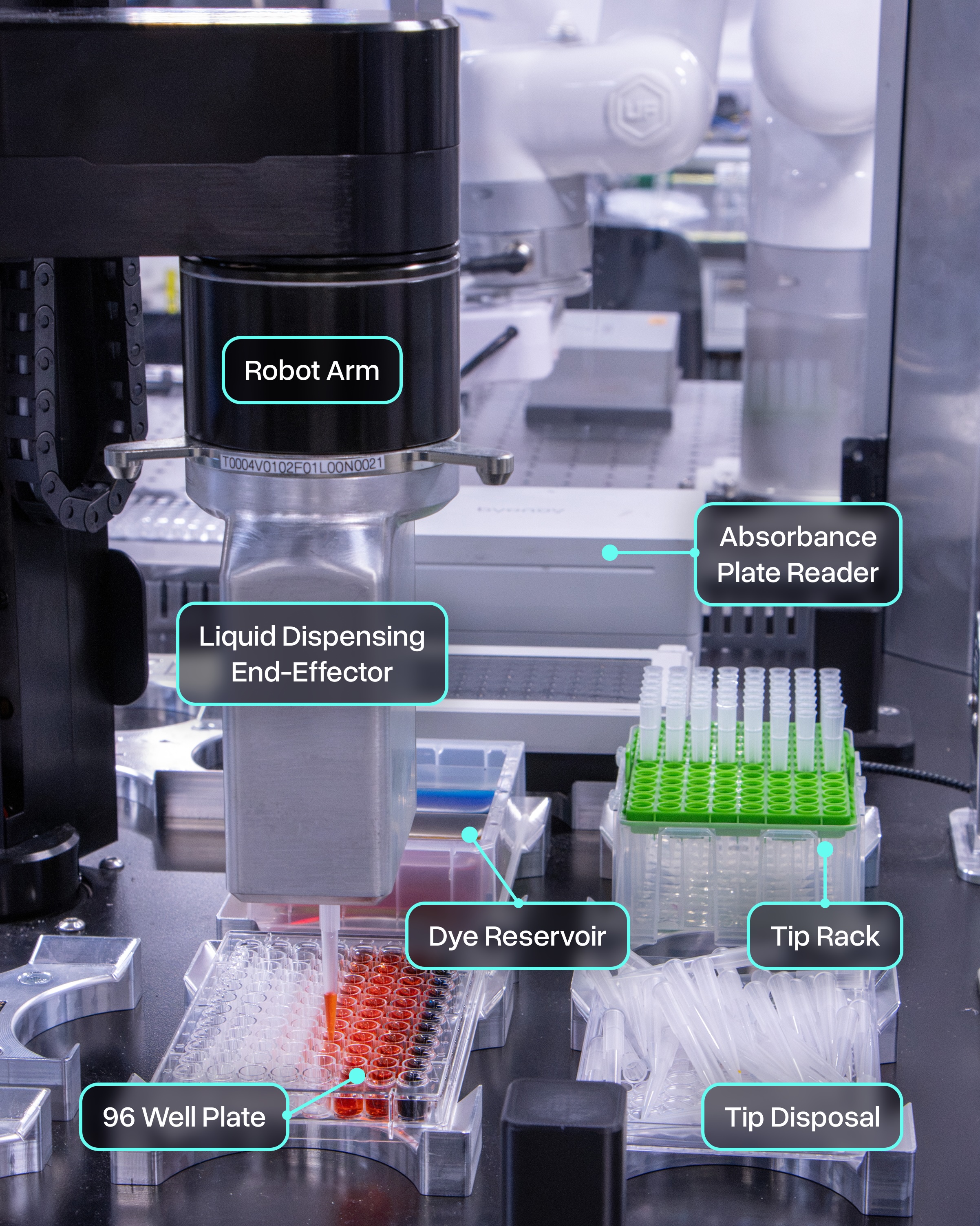}
    \caption{The experimental setup within the Trilobot robot platform includes a Byonoy Absorbance 96 Automate plate reader, a standard 96-well plate, a rack of 96 pipette tips, a 4-well dye reservoir, and a bin for used pipette tips.}
    \label{fig:setup}
\end{figure}

\begin{figure*}
    \centering
    \includegraphics[width=\textwidth]{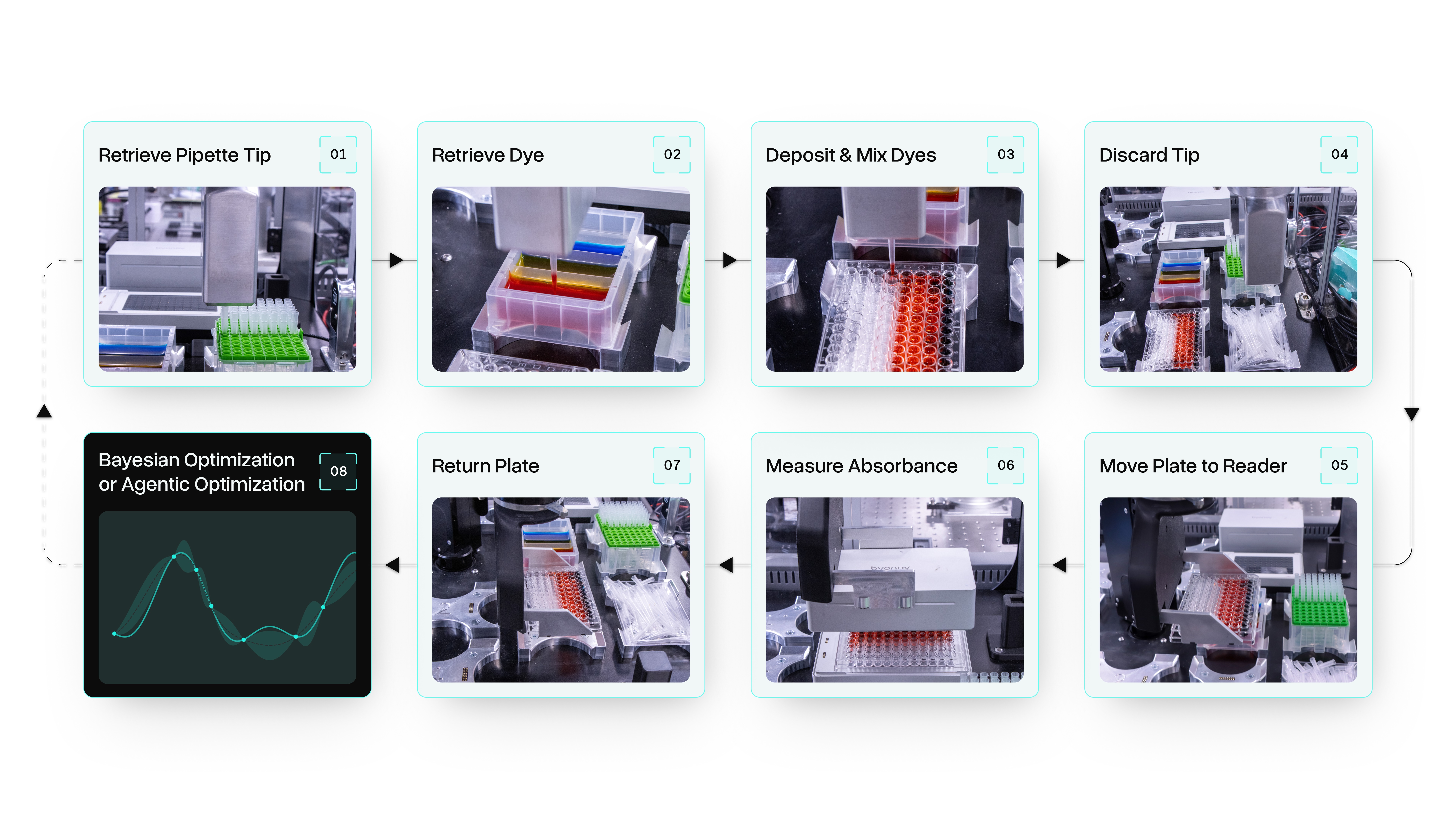}
    \caption{The autonomous workflow. Each iteration, the decision-making algorithm (BO or the LLM agent) proposes a candidate mixture, automating the decision-making process. The robotic system produces this candidate and measures its absorbance, automating the physical execution process. These data are used to update the model, which then generates the next candidate mixture.}
    \label{fig:workflow}
\end{figure*}

Our physical testbed for an SDL is built around a general-purpose laboratory robot (Trilobot, Trilobio) enclosed in a transparent housing, with interchangeable end effectors for pipetting and gripping (Fig.~\ref{fig:trilobot}). This modular design allows a single robot to handle a range of laboratory objects, including microplates, pipette tips, and measurement instruments, without reconfiguring the surrounding workcell. Within the Trilobot, our workspace is arranged with five fixed elements: a 96-well microplate, an absorbance plate reader, a set of dye reservoirs, a tip rack from which the robot retrieves a new pipette tip each iteration, and a bin for used pipette tips (Fig.~\ref{fig:setup}). The robotic platform performs the pipetting and labware manipulation required to formulate each experiment, while the plate reader measures the resulting liquid formulations. The resulting data are then used by an AI-based decision-making algorithm to design subsequent experiments without human intervention (Fig.~\ref{fig:workflow}). This testbed supports a wide range of experiments based on low-viscosity liquid dispensing and filter-based absorbance at three wavelengths as the primary performance measurement. Together, this hardware configuration establishes the physical foundation for laboratory autonomy, which enables the coupling of flexible, commercially available instrumentation with the ability to operate with closed-loop, unattended operation. 

We evaluate the integrated platform on a color formulation task that involves identifying the volumes of four liquids (a red dye solution, a yellow dye solution, a blue dye solution, and water) that, when mixed, reproduce a target absorbance spectrum. The three dye solutions are prepared by adding three drops of grocery-store food coloring to water. The robot dispenses all four liquids into a single well of the 96-well microplate under the constraint that their volumes sum to a fixed total of 250~µL, then homogenizes the mixture by repeated aspiration and dispensing within the well. The absorbance spectrum of each well is then measured with an absorbance microplate reader, which reads all 96 wells of a microplate over a photometric range of 0.0 to 4.0 OD. 

To quantitatively characterize each mixture, we record absorbance at three wavelengths (400~nm, 600~nm, and 700~nm), set by the plate reader's installed filters. These three wavelengths do not uniquely determine a mixture's full spectrum, since metamers can share absorbance at the measured wavelengths while differing at others; we nonetheless use this three-wavelength representation determined by our existing instrumentation. We use the device measurement accuracy specifications to determine if a target spectrum has been achieved within the specified measurement error (400~nm: $\leq$ 1.5\% + 0.010~OD from 0.0–2.0~OD; $\leq$ 3.0\% + 0.010~OD from 2.0–3.0~OD; $\geq$ 450~nm: $\leq$ 1.0\% + 0.010~OD from 0.0–2.0~OD; $\leq$ 1.5\% + 0.010~OD from 2.0–3.0~OD).

We formulate our optimization problem as follows. Let $\boldsymbol{x} = (x_1, x_2, x_3, x_4)$ be the volumes of the four liquids and $\boldsymbol{y} = (y_1, y_2, y_3)$ be the target absorbances at 400~nm, 600~nm, and 700~nm. Let $S(\boldsymbol{x})$ represent the physical process of mixing and measuring that results in the measured absorbance spectrum for mixture $\boldsymbol{x}$. Our optimization problem is then:
\begin{equation}
\min_{\boldsymbol{x}}{\ {\left\lVert S\left(\boldsymbol{x}\right)-\boldsymbol{y}\right\rVert }_{2}}
\text{ s.t. }\sum _{i=1}^{4} x_{i}=250\text{\ and\ }x_{i}\geq 0
\end{equation}

\subsection{Bayesian Optimization}
To demonstrate the capability of our autonomous lab to solve black-box optimization problems, we use Bayesian optimization to match a desired absorbance spectrum. Dye mixtures are selected using BO implemented with the BoTorch library \cite{Balandat2020_BOTORCH}. We fit three independent Gaussian processes to model the absorbance at each of the three measured wavelengths as a function of the input volumes. To handle the constraint that the input volumes sum to a constant value, we reduce the number of input parameters to the three dyes with the constraint that their sum is no greater than 250. The water volume is set to the remaining volume.
\begin{figure*}
    \centering
    \includegraphics[width=\textwidth]{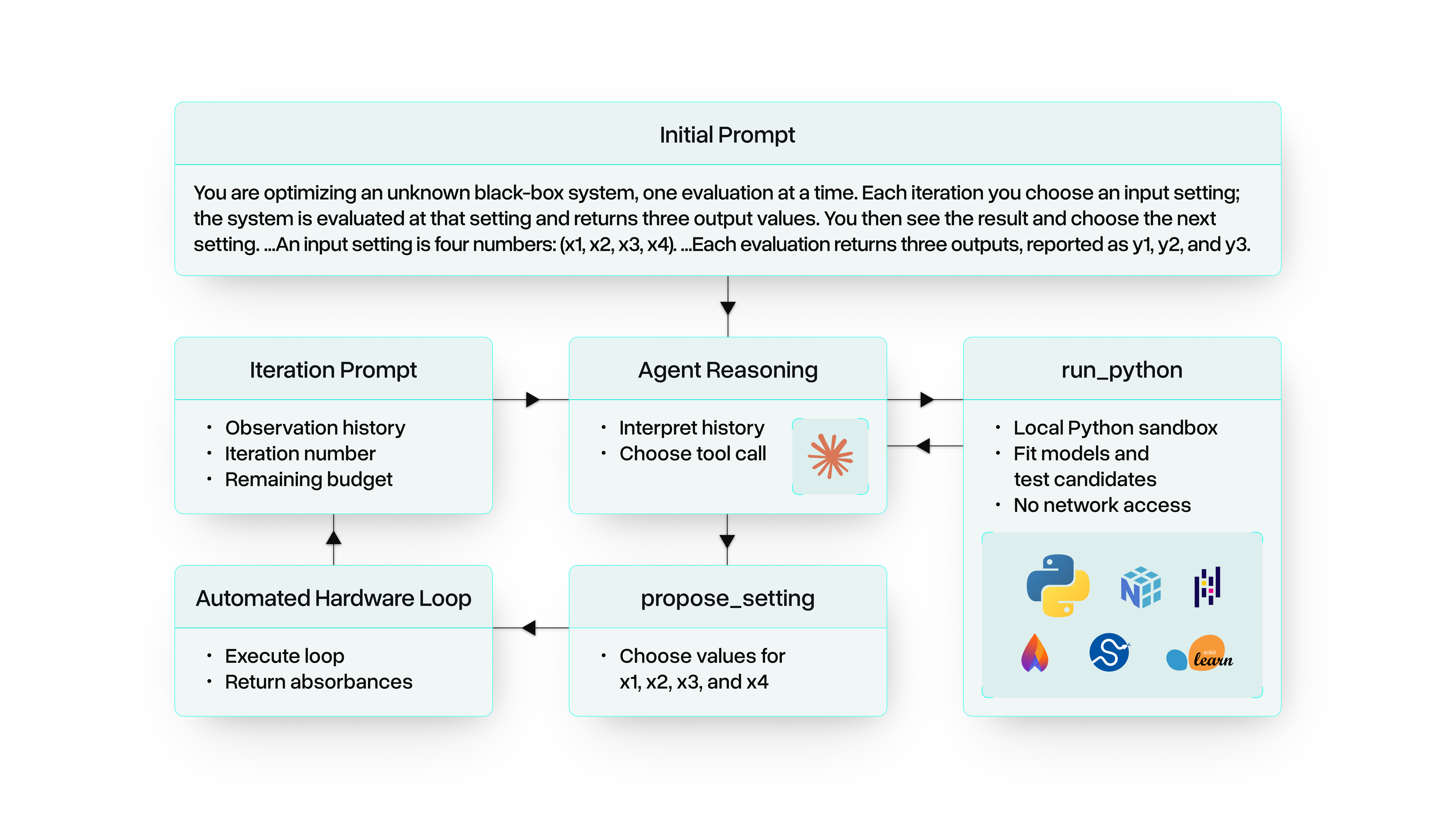}
    \caption{The agent harness for design optimization. The agent is first given the initial prompt containing the general instructions and optimization problem framework. During each iteration, the agent reasons and can select from two client tools: \texttt{run\_python} and \texttt{propose\_setting}. The agent can choose to run Python code locally, as many times as needed, to fit models before deciding on the next setting, which the automated hardware loop evaluates. The measured absorbance values are appended to the observation history, which is given as the next prompt to the agent in a continuous context.}
    \label{fig:agentic}
\end{figure*} 

Each BO campaign begins with 12 mixtures chosen by random sampling within the feasible, sum-constrained input space, seeding the Gaussian processes with an initial dataset. The following 36 mixtures are chosen iteratively, one at a time, using the \texttt{qLogExpectedImprovement} acquisition function \cite{Ament2023_Unexpected}. The objective for the BO is the negative mean squared error between the three outputs of our model and the target. The optimization hyperparameters are listed in Supplementary Information~\ref{app:bo}. 

\begin{table}[htbp]
  \centering
  \caption{Configuration of the closed-loop color-formulation campaigns.}
  \label{tab:campaign_config}
  \begin{tabularx}{\linewidth}{>{\raggedright\arraybackslash}p{3cm} X}
    \toprule
    Parameter & Value \\
    \midrule
    \multirow{4}{=}{Input variables}
      & $x_{1}$: red dye volume ($\mu$L) \\
      & $x_{2}$: yellow dye volume ($\mu$L) \\
      & $x_{3}$: blue dye volume ($\mu$L) \\
      & $x_{4}$: water volume ($\mu$L) \\
    \midrule
    Constraint & $x_{1}+x_{2}+x_{3}+x_{4}=250~\mu\mathrm{L}$ \\
    \midrule
    \multirow{3}{=}{Measured outputs}
      & $y_{1}$: absorbance at 400~nm (OD) \\
      & $y_{2}$: absorbance at 600~nm (OD) \\
      & $y_{3}$: absorbance at 700~nm (OD) \\
    \midrule
    Total iterations per campaign & 48 \\
    \midrule
    Wall-clock time per campaign & $\sim$4~h, fully autonomous \\
    \midrule
    Number of campaigns & 6 \\
    \bottomrule
  \end{tabularx}
\end{table}
\subsection{Agentic Optimization}
For comparison with traditional approaches such as BO, we created a custom agent harness to solve the color formulation optimization problem (Fig.~\ref{fig:agentic}). At each iteration, the agent is given the observation history of inputs and outputs to the optimization problem, and it selects the next mixture to test through reasoning and code execution.

We use the Claude Opus 5 model with thinking effort set to ``high'' through the Claude Messages API platform, sending text prompts and receiving generated responses. We maintain a single continuous conversation for an optimization campaign of 48 iterations so that its entire optimization process remains in context. The agent then uses the history of the entire conversation to select the candidate inputs to be created for the next iteration, optionally executing Python code in a local sandbox to inform its decisions. We save the responses from Opus 5, which include the rationale for selecting each candidate mixture.

The optimization problem is presented to the agent as a black-box problem, i.e., the prompt uses the variable names x and y for the input and output, so the agent is unaware that the problem is color formulation. This task-agnostic prompt ensures that the agent does not rely on its prior knowledge of color theory to approach the optimization problem, as we wish to demonstrate that our framework can apply to a broader range of experimental workflows. 

We give the agent access to a Python sandbox with NumPy, SciPy, pandas, scikit-learn, and BoTorch installed. The agent can call a tool we define, \texttt{run\_python}, by sending a generated code string that is executed in this local Python sandbox. The agent does not have access to any external tools, such as web search. We impose these limitations to demonstrate the agent's ability to solve our black-box optimization problem with basic machine learning toolkits. Furthermore, this sandbox isolation prevents access to the experimental data from previous campaigns present in our code repository, which would give the agent additional data to aid its optimization task. This restriction ensures that there is a fair comparison against the Bayesian optimization approach.

\section{Autonomous Experimental Campaigns}
We ran six campaigns, three with the Bayesian optimizer and three with the agentic optimizer. Each campaign consisted of 48 iterations completed autonomously in approximately four hours without any manual intervention (Table~\ref{tab:campaign_config}). We generated three target absorbance spectra by creating three random mixtures and measuring their absorbance spectra, and we ran independent campaigns on each target using the Bayesian optimization method and the agentic optimization method. Our closed-loop workflow attempted to reproduce the target spectra by iteratively selecting, dispensing, and measuring new dye mixtures. Across the six campaigns, our system achieved an average RMSE of $9.95 \times 10^{-3}$ OD, indicating that the platform consistently converged toward its target regardless of which spectrum it was assigned to match.

We used the Absorbance 96 Automate reader’s measurement accuracy specifications to determine if a measured absorbance spectrum matches a target within the specified accuracy range. We then computed the number of iterations each campaign takes to match the target absorbance spectrum within the measurement error (Table~\ref{tab:iterations_to_target}), and we found that all but one campaign matched the target within the error band. The agentic optimizer matched the target spectra in fewer iterations and succeeded on target 3, whereas the Bayesian optimizer failed to match that target within 48 iterations.

\begin{figure}
    \centering
    \includegraphics[width=\linewidth]{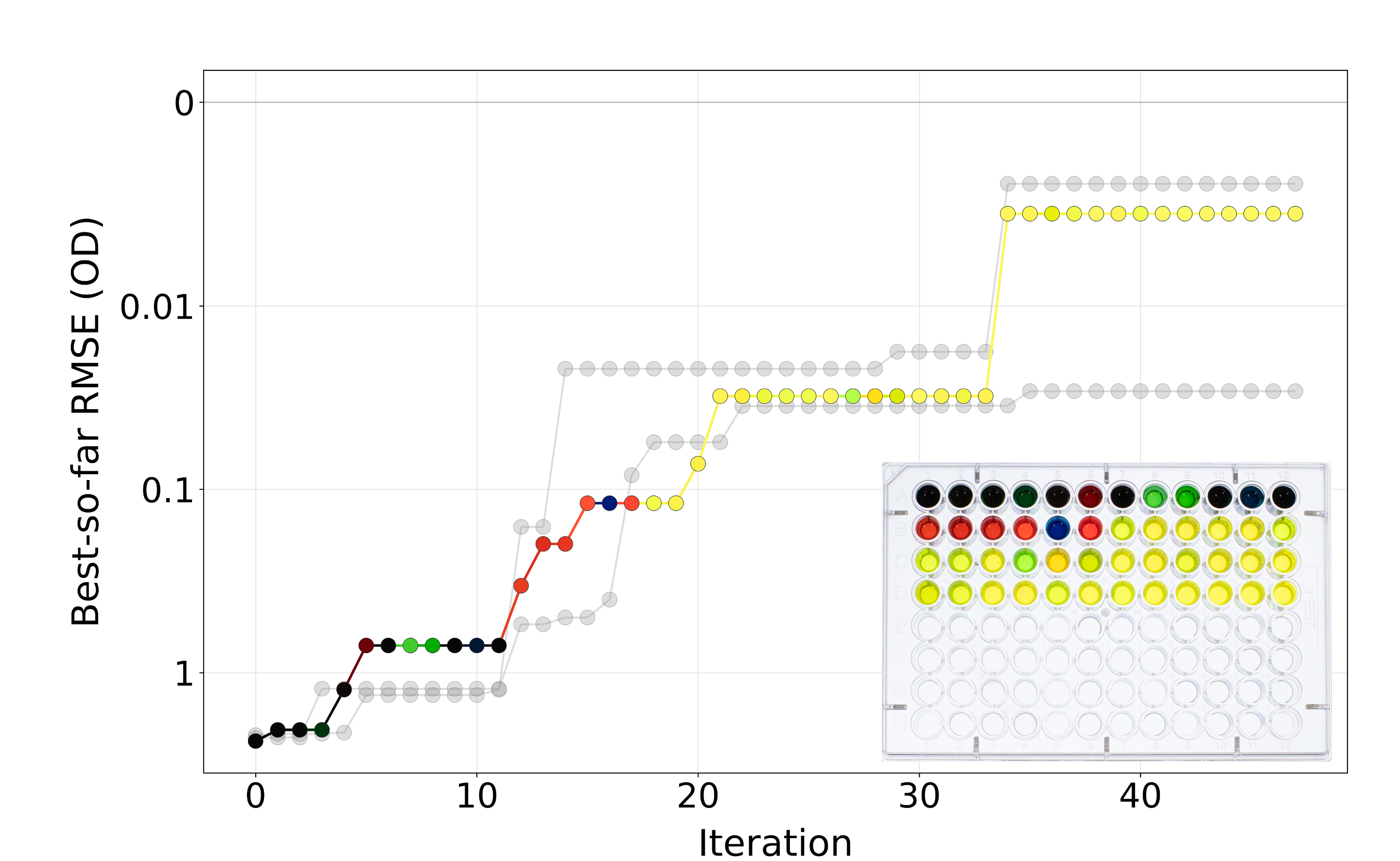}
    \caption{The best-so-far RMSE per iteration of the three BO campaigns is shown, with one campaign featured in color with its microplate. The y-axis is a symmetric logarithmic axis (logarithmic above 0.01~OD, linear between 0 and 0.01~OD), where higher is better. The marker colors show the photographed color of the mixture measured at that iteration. The first row of the microplate contains the 12 randomly generated mixtures, and the next three rows contain the 36 mixtures generated by the Bayesian optimizer. The campaign shown achieved an RMSE of $5.47 \times 10^{-3}$~OD.}
    \label{fig:bo}
\end{figure}

The first campaign with Bayesian optimization illustrates convergence to the target in detail (Fig.~\ref{fig:bo}). During the 12 initial random iterations, the error achieved by the best mixture so far improves as random mixtures happen to match the target spectrum more closely. Once the Bayesian optimizer takes over, the best-so-far trace climbs steadily through the following iterations and approaches low error by iteration 35, after which additional iterations produce no further improvement. The inset microplate image makes this progression visible directly, as the four rows of wells shift from the dark, heterogeneous mixtures of random exploration to the visually consistent yellow mixtures that dominate the final iterations, matching the target spectrum assigned to this campaign. 

\begin{figure*}
    \centering
    \includegraphics[width=\textwidth]{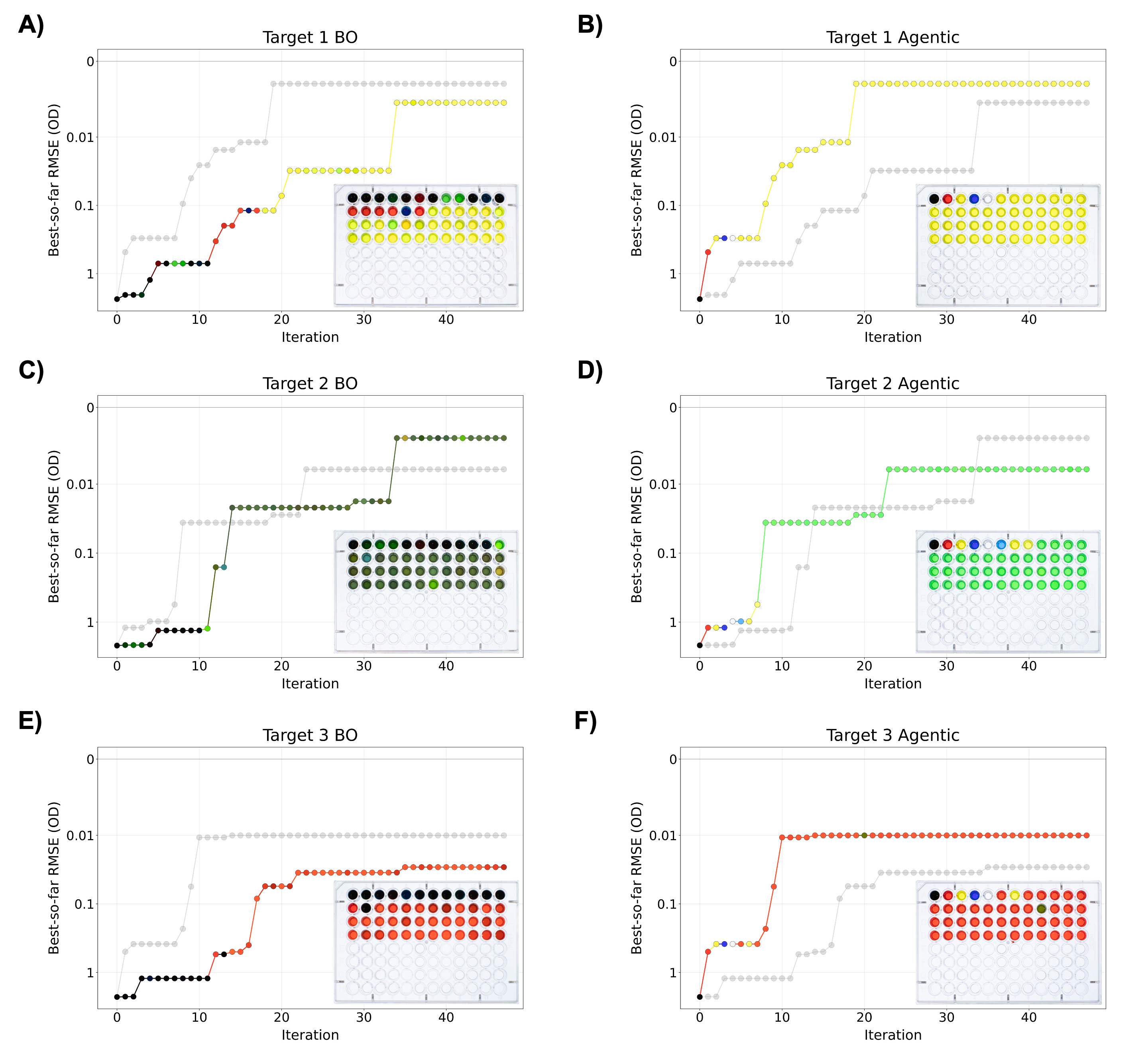}
    \caption{Best-so-far RMSE per iteration for all six campaigns. The best RMSE achieved by the end of each campaign are (A) $5.47 \times 10^{-3}$~OD, (B) $2.98 \times 10^{-3}$~OD, (C) $4.00 \times 10^{-3}$~OD, (D) $8.08 \times 10^{-3}$~OD, (E) $2.92 \times 10^{-2}$~OD, and (F) $1.00 \times 10^{-2}$~OD. Note that for target 2, the resulting colors from the Bayesian optimizer and the agentic optimizer are visually different, though they achieve similar absorbances at the measured wavelengths of 400~nm, 600~nm, and 700~nm.}
    \label{fig:results}
\end{figure*}

The other campaigns followed the same overall pattern but converged to different final accuracies (Fig.~\ref{fig:results}). The second campaign of the Bayesian optimizer (Fig.~\ref{fig:results}C) reached its best-so-far plateau at iteration 34 and achieved an RMSE of $4.00 \times 10^{-3}$ OD, close to that of the first campaign, with its microplate showing a corresponding shift from black to a range of green and olive mixtures. The third campaign of the Bayesian optimizer (Fig.~\ref{fig:results}E) also converged well below the level of unguided sampling but plateaued at a higher RMSE of $2.92 \times 10^{-2}$ OD, roughly five-fold larger than that of the first campaign. Its microplate shows the later rows converging almost uniformly to red, consistent with the optimizer settling into a narrow region of the input space rather than continuing to resolve finer differences in mixture composition.

\begin{table}[htbp]
  \centering
  \caption{The number of iterations to reach the target within the measurement error band for each campaign.}
  \label{tab:iterations_to_target}
  \begin{tabular}{l c >{\centering\arraybackslash}p{5cm}}
    \toprule
    Method & Target & Number of iterations to reach target within error band \\
    \midrule
    \multirow{3}{*}{BO}
      & 1 & 21 \\
      & 2 & 34 \\
      & 3 & --- \\
    \midrule
    \multirow{3}{*}{Agentic}
      & 1 & 10 \\
      & 2 & 23 \\
      & 3 & 11 \\
    \bottomrule
  \end{tabular}
\end{table}

In the three campaigns with the agentic optimizer (Fig.~\ref{fig:results}B, D, F), Opus 5 chose the candidate mixtures through reasoning and Python code execution. In all three campaigns, the agent decided to begin by sampling the center of the design space (equal volumes for each of the four liquids), followed by the four vertices of the design space simplex (250~µL of a single liquid). Following this, the agent created models, made local refinements, and then repeated similar samples once it determined the remaining error was due to noise. 

The agent was able to adjust its approach and modeling choices during the campaigns, employing several different methods to pursue a lower error. The agent used Python to fit linear models, Hill curves, power law response curves, and Michaelis–Menten curves. It used strategies such as recency weighting, checking for drift, and computing noise correlation between output channels. For target 3, the agent used Monte Carlo simulations to determine how to sample to achieve marginally better results once it concluded it was at the noise floor, though this failed to reduce the error. Although scikit-learn and BoTorch were available, the agent never chose to use Bayesian optimization. The details of the agentic optimization runs are given in Supplementary Information~\ref{app:run}.

The agentic optimizer eventually shifted from exploration to exploitation once it determined that most of the variance in the measured spectra was due to noise. It made progressively smaller adjustments to the proposed mixtures to further reduce the error. This matched the behavior of the Bayesian optimizer, which, in its campaigns, exploited narrow regions of the input space in later iterations.

The agentic optimizer required fewer iterations to match the target than traditional Bayesian optimization across all three targets, achieving a 52\% reduction for target 1 and a 32\% reduction for target 2, and converging on a formulation within 11 iterations for target 3, a target for which Bayesian optimization failed to reach the error band within 48 iterations (Table~\ref{tab:iterations_to_target}). These results suggest that agentic optimization may be a more effective approach for our color formulation task and may also reduce the implementation burden associated with traditional decision-making methods. Unlike BO, which requires custom optimization code, hyperparameter selection, and a pre-specified model structure, the agentic optimizer selects and implements its own models.

In terms of time, the Bayesian optimizer and agentic optimizer were comparable. The agentic optimizer took an average of 19 seconds to produce a candidate mixture each iteration, whereas the Bayesian optimizer took an average of 16 seconds. The time during each iteration was predominantly spent on the physical automation loop, which took approximately 4.5 minutes per iteration. The total API cost across the three campaigns using the agentic optimizer was \$20.83. 

Taken together, these six campaigns demonstrate that the closed-loop workflow converges on previously unseen targets, and that the shape of this convergence, a period of initial exploration followed by sustained improvement, replicates across independent runs rather than depending on a particular target spectrum. Each campaign ran to completion, 48 iterations and roughly four hours of autonomous dispensing, measurement, and re-optimization, without a single manual intervention, providing six independent demonstrations that the testbed's closed loop operates as intended under ordinary laboratory conditions. Furthermore, we highlight the benefit of using an agentic approach, which exceeded the performance of our Bayesian optimization approach while monitoring the experiments to choose different models appropriate to the incoming data during a campaign.

\section{Discussion}
The SDL community has expanded considerably in recent years, broadening both the hardware and software infrastructure available and the scientific domains to which SDL technology is applied. Parallel efforts have sought to advance SDL technology by developing testbeds that serve as a basis for prototyping hardware and software. Such testbeds provide a means of learning about SDL operation, integrating and evaluating hardware, and benchmarking decision-making algorithms. Although the development of testbeds offers a significant opportunity to assess and mature SDL technology, comparatively little attention has been paid to the role such testbeds play in SDL technology development, and much of the available evidence instead centers on bespoke, single-instance platforms built to address specific scientific applications.

Here, we present a testbed assembled from general-purpose, commercially available components that offers a shared reference point for evaluating decision-making strategies across research groups. The testbed directly addresses the challenge established at the outset: evaluating decision-making approaches within a common platform built from off-the-shelf instrumentation, rather than custom hardware. Notably, this choice carries a cost trade-off. Assembling the testbed required substantial investment, reflecting the reliance on commercially available hardware from vendors rather than custom-built or lower-cost alternatives. The use of these instruments also reflects recent industry progress in developing hardware suited for SDL integration, including the robotic platform and the absorbance plate reader used in this study. These instruments now expose Python APIs, enabling closed-loop operation and integration. We do not claim that this platform is the optimal physical testbed available, only that it constitutes a commercially available reference system. Its design illustrates how recent advances in commercially available hardware can support the maturation of SDL technology. As a result, the availability of the physical testbed enables comparison of decision-making algorithms on the color-formulation problem, a task previously used to demonstrate the operating principles of SDL technology. Importantly, this progress in the SDL field depends on continued progress in software and decision-making layers coupled with advances in hardware.

Using the testbed described above, we evaluated two decision-making algorithms across six independent campaigns. The platform operated without any manual correction for a combined total exceeding 24 hours, roughly four hours per campaign. During these campaigns, RMSE fell from above 1 OD, characteristic of the initial random sampling phase, to an average of $9.95 \times 10^{-3}$ OD, achieved through autonomous selection of successive experiments. These results provide evidence that an autonomous decision-making layer can identify and exploit structure in an experimental space over extended, unattended operation, a capability that automation alone does not confer. 

The Bayesian optimizer succeeded in achieving an absorbance spectrum within the measurement error of the target spectrum for two of three targets, and the agentic optimizer succeeded for all three targets. These results further suggest that a fully agentic optimization method can be equally or more effective for proposing candidate mixtures to match a target spectrum, though further experimentation may be required to reach a more confident conclusion. The agentic optimizer found a match in 52\% fewer iterations for target 1, 32\% fewer iterations for target 2, and it succeeded on target 3 where BO failed to match the target within 48 iterations. The Bayesian optimization method required custom BoTorch code and manual hyperparameter selection. The agentic optimization method subsumes these steps, selecting and implementing models to propose candidate mixtures, and updating these models during a campaign. 

The low-risk nature of the color-formulation task is central to the broader significance of these results. This task allows newer commercial hardware and decision-making strategies alike to be evaluated and iterated upon without the material or safety consequences that a domain-specific application would carry. The platform provides a reproducible, low-stakes setting in which the operational reliability and decision-making layer of an SDL can be assessed on equal footing.

\section{Conclusion}
In this work, we have presented a physical testbed for self-driving laboratories, demonstrated through a minimal closed-loop discovery task, the autonomous formulation of dye mixtures to match target absorbance spectra. Across six independent campaigns totaling more than 24 hours of unattended operation, the system completed 48 closed-loop iterations per campaign, selecting every mixture, dispensing every volume, and taking every measurement without human intervention. Root mean squared error fell from above 1 OD during initial random sampling to an average of $9.95 \times 10^{-3}$ OD by the end of optimization, a reduction achieved through autonomous decision-making. Within this same closed loop, substituting a large language model for a Bayesian optimization algorithm indicated that an agentic decision-making mechanism can match target absorbance spectra in fewer iterations than traditional Bayesian optimization. Taken together, these results indicate that a commercial laboratory robot and measurement instrument, coupled with either an open-source BO library or a commercial LLM API through their existing programmatic interfaces, are sufficient to execute end-to-end autonomous experimentation, supporting the evaluation and continued development of physical testbeds for SDL technology.

We regard this system as a foundation rather than a finished platform. Extending the system along the automation axis will involve coupling additional analytical instruments to the existing plate reader and expanding from single-well to batched proposal and measurement, increasing both the throughput and the informational content of each closed-loop cycle. Extending the agentic control of the SDL will involve exposing the Trilobot API to the agent to allow it to generate adaptive physical workflows. Natural next steps include applying this system to problems of direct practical relevance, including buffer, excipient, and reaction-condition optimization. Because the unit operations demonstrated here generalize well beyond color formulation, the same physical platform is, in principle, already positioned to support such extensions.

\section*{Acknowledgements}
This work was funded under the LLNL Laboratory Directed Research and Development project 25-ERD-032. This work was performed under the auspices of the U.S. Department of Energy by Lawrence Livermore National Laboratory under Contract DE-AC52-07NA27344. LLNL-JRNL-2025156.
\printcredits

\section*{Declaration of Competing Interests}
M.S. and R.A.-N. are co-founders and employees of Trilobio, a company that develops commercial automated laboratory platforms for biological experiments. The remaining authors declare that they have no competing interests.

% \bibliographystyle{cas-model2-names}
\bibliographystyle{cas-model2-names-unsrt}

\bibliography{refs}


\clearpage

\appendix
\section*{Supplementary Information}
\section{BO Hyperparameters}
\label{app:bo}
\begin{tabular}{ll}
\toprule
\textbf{Parameter} & \textbf{Setting} \\
\midrule
Kernel         & RBF (default) \\
Noise model    & GaussianLikelihood (default) \\
\texttt{num\_restarts} & 32 \\
\texttt{raw\_samples}  & 512 \\
\texttt{batch\_limit}  & 5 \\
\texttt{max\_iter}     & 200 \\
\bottomrule
\end{tabular}

\section{Agentic Optimization Run Details}
\label{app:run}
For all three campaigns, the agent started with a mixture of equal values for all four variables (62.5). It then tried each of the four inputs separately with a value of 250. All three campaigns then followed the same pattern of fitted models, local refinements, and repeated near-duplicate candidates to achieve a better score from favorable noise draws. For each iteration, the agent had the option to use Python code to inform its next candidate proposal. If the agent did not use Python, it reasoned without code execution or followed a plan from previous iterations.

During its campaign for target 1, the agentic optimizer called the \texttt{run\_python} tool 12 times. It began with the 5-well survey (the center point plus the four vertices). Then, it tried to reason about how $x_{2}$ influenced $y_{1}$ through sampling a few wells, since it determined from its initial 5-well survey that $x_{2}$ was the most important input to achieve the target. Its first Python call was to print a table of the data in iteration 11. In iterations 13--19, the agent made multiple Python calls to fit linear models on a subset of the data close to the previous proposals. After this, it concluded that it was at the noise floor, so it planned to repeat a candidate input for multiple iterations. Its best result was achieved in iteration 20, and the subsequent iterations repeated similar inputs. In iterations 24--32, the agent called Python to do a drift check with a linear model, and it also used Python to fit a Hill curve. In iterations 33--35, the agent called Python to perform recency-weighted estimates, a drift check, and a noise correlation. In the remaining iterations, the agent repeated near-duplicate candidates.

During its campaign for target 2, the agentic optimizer called the \texttt{run\_python} tool 13 times. It began with the 5-well survey. It determined that $x_{1}$ was unimportant; it then used the next iterations to explore the effects of $x_{2}$ and $x_{3}$. Its first Python call was on iteration 9, where it fit two curves, a saturating exponential and a Hill curve. In the following iterations, the agent reasoned through small adjustments by performing calculations itself without Python. In iterations 14--16, the agent called Python to fit a power law curve and a straight line. In iterations 17--24, the agent made more Python calls to perform linear regressions with data points close to the previous candidates. The best result was achieved on iteration 24. The agent continued with local linear models, and it reasoned without code that there might be a correlated noise factor. In iterations 38--48, the agent repeated near-identical candidates.

During its campaign for target 3, the agentic optimizer called the \texttt{run\_python} tool 33 times. After its initial 5-well survey, it called Python on iteration 6 to fit a linear model, which did not lead to a close match with the target due to saturation on $y_{1}$. It then used subsequent iterations to try small values of $x_{1}$ and $x_{2}$. In iterations 9--11, the agent called Python multiple times to fit a Michaelis-Menten curve, Hill curves, and a linear fit. One of these Python calls crashed due to a missing import, which caused the agent to call the Python tool again with corrected code. It then searched a fitted Hill curve model for the lowest predicted error. In iterations 12--15, the agent used Python to fit more Hill curves and a straight line on recent well data. It achieved its best score on iteration 15. In iterations 16--31, the agent fit more curves and straight lines, and it determined that it was at the noise floor. The agent called Python to perform a Monte Carlo search to determine how to optimize for lucky noise draws. It also used Python to perform a drift check. In iterations 32--38, it used Python to write notes on its model and its plan to a text file. In iterations 39--44, it used Python to run more Monte Carlo simulations and drift checks. In the remaining iterations, it repeated the same candidate.
\end{document}